\documentclass[conference]{IEEEtran}

\usepackage[super]{nth}
\usepackage{hyperref}
\usepackage{cite}
\usepackage[pdftex]{graphicx}
\usepackage{subcaption}
\usepackage{balance}
\usepackage{booktabs} 
\usepackage[table]{xcolor}

\usepackage{amsmath,amsfonts}
\usepackage[ruled,linesnumbered]{algorithm2e}
\usepackage{tablefootnote}
\usepackage{amssymb}
\usepackage{pifont}
\newcommand{\cmark}{\ding{51}}%
\newcommand{\xmark}{\ding{55}}%

\graphicspath{ {./images/} }

\newcommand{\Reconv}{\texttt{Reconvene }}

\newcommand{\Reconvv}{\texttt{Reconvene}}

\begin{document}
\include{pythonlisting}

\title{Rapid Deployment of DNNs for Edge Computing via Structured Pruning at Initialization\vspace{-0.35em}}

\author{
\IEEEauthorblockN{ }
    \IEEEauthorblockN{Bailey J. Eccles\IEEEauthorrefmark{1}, Leon Wong\IEEEauthorrefmark{2}, and Blesson Varghese\IEEEauthorrefmark{1}}
    
    \IEEEauthorblockA{\IEEEauthorrefmark{1}\textit{School of Computer Science, 
    University of St Andrews, UK}}
    
    \IEEEauthorblockA{\IEEEauthorrefmark{2}\textit{Autonomous Networking Research \& Innovation Department, Rakuten Mobile, Inc., Japan}
    \\
    Corresponding author: \texttt{bje1@st-andrews.ac.uk}}
    \\[-15pt]
}

\maketitle
\thispagestyle{plain}
\pagestyle{plain}

\begin{abstract}
Edge machine learning (ML) enables localized processing of data on devices and is underpinned by deep neural networks (DNNs). However, DNNs cannot be easily run on devices due to their substantial computing, memory and energy requirements for delivering performance that is comparable to cloud-based ML. Therefore, model compression techniques, such as pruning, have been considered. Existing pruning methods are problematic for edge ML since they: (1) Create compressed models that have limited runtime performance benefits (using unstructured pruning) or compromise the final model accuracy (using structured pruning), and (2) Require substantial compute resources and time for identifying a suitable compressed DNN model (using neural architecture search). In this paper, we explore a new avenue, referred to as Pruning-at-Initialization (PaI), using structured pruning to mitigate the above problems. We develop \Reconvv, a system for rapidly generating pruned models suited for edge deployments using structured PaI. \Reconv systematically identifies and prunes DNN convolution layers that are least sensitive to structured pruning. \Reconv rapidly creates pruned DNNs within seconds that are up to 16.21$\times$ smaller and 2$\times$ faster while maintaining the same accuracy as an unstructured PaI counterpart.
\end{abstract}

\begin{IEEEkeywords}
Deep Neural Networks, Edge Computing, Model Compression, Structured Pruning
\end{IEEEkeywords}

\IEEEpeerreviewmaketitle

\section{Introduction}
\label{sec:introduction}
Deep neural networks (DNNs) are used in many applications to process and analyze data at the network edge for mitigating the challenges in sending all the data to the cloud~\cite{edge}. For instance, security cameras for facial recognition~\cite{camera} and wearable health monitors~\cite{wearable} benefit from edge computing. The DNN models used in these settings are often over-parameterized for the application task, requiring a large amount of computing resources for training and inference~\cite{han2015, pruning_survey}.

Embedded and mobile edge devices cannot support large cloud-based DNN models due to computational, memory, and energy constraints~\cite{intro_rc}. Therefore, model compression methods that reduce the resource requirements of training and inference while preserving task accuracy are used~\cite{model_comp_survey}. Compression methods include quantization~\cite{quant}, knowledge distillation~\cite{kd}, neural architecture search~\cite{ofa}, and pruning~\cite{he2017, han2015}.

Model pruning removes specific parameters from over-parameterized and dense DNNs while tailoring models for specialized tasks. In contrast to other model compression methods, model pruning is beneficial in edge computing environments, where optimizing models is required for diverse applications with heterogeneous computational constraints and capabilities~\cite{ofa, ECCLES2023}. For example, edge-based model training paradigms, such as federated learning, make use of model pruning to expedite the training time of straggler devices~\cite{prunefl, zerofl}. Model pruning is categorized as unstructured pruning (UP) and structured pruning (SP). UP sets parameter weights to zero, while SP removes groups of parameters. 

\begin{figure}[!t]
\centerline{\hspace*{0.5cm}\includegraphics[width=0.49\textwidth]{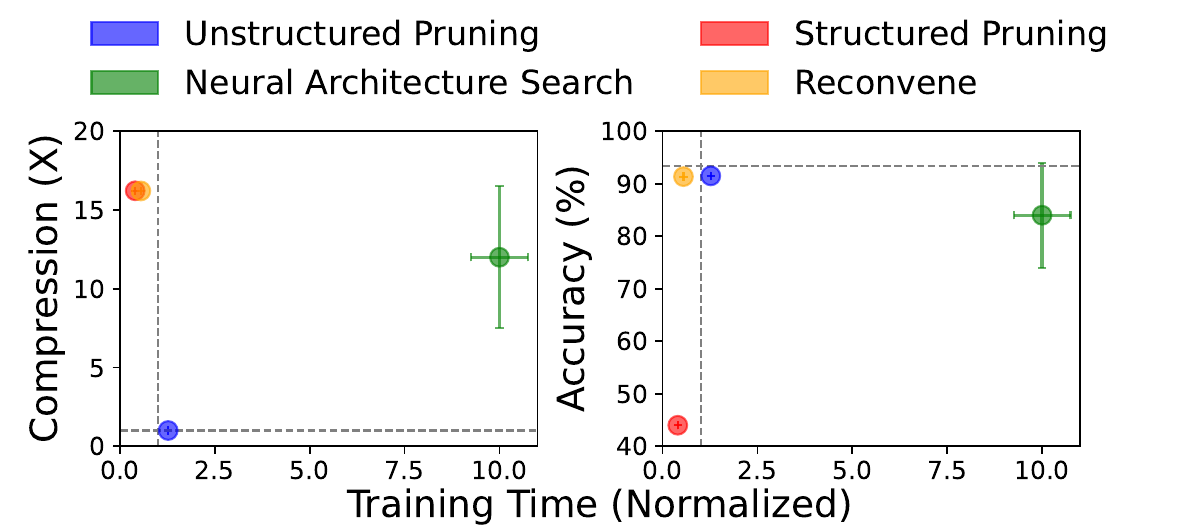}}
\caption{Evaluating model compression methods to reduce parameter count of VGG-16 (on the CIFAR-10 dataset) by 50$\times$. Dashed lines are the baseline values of an uncompressed dense VGG-16. The bar for neural architecture search includes the discovery time for generating a range of compressed models with different levels of compression and accuracy.}
\vspace{-10pt}
\label{fig1}
\end{figure}

Figure~\ref{fig1} shows runtime differences between compression methods. UP retains model accuracy, whereas SP enhances compression and training speed. Compared to pruning, neural architecture search (NAS) finds a range of models from a vast search space but takes longer for searching and training~\cite{ofa}.

Pruning and NAS offer many benefits for edge computing. However, there is a three-fold challenge that impacts the deployment of compressed models in the edge setting: (1) Retaining the accuracy of the compressed model similar to that of the original dense model, (2) Achieving model compression that empirically decreases training and inference latency and model size, and (3) Discovering a pruned model rapidly and efficiently. While existing methods can address up to two of these challenges simultaneously, they do not address all three at the same time (highlighted in Table~\ref{tab:comp}). For example, models generated by SP are smaller, faster, and easily discoverable but often have low accuracies, and therefore, lack usability for accuracy-critical edge applications. An ideal system for pruning will be underpinned by a method that addresses all three challenges considered above. 

This paper introduces \Reconvv, a system that addresses the above three challenges for pruning large DNN models to create compressed models suitable for resource-constrained devices in edge computing environments. \Reconv achieves this by using a novel combination of both unstructured and structured pruning methods at model initialization (PaI; i.e., before training) and is the first system to maintain model accuracy up to extreme levels of model compression in this category~\cite{frankle2021pruning, cai2022structured}. Other pruning systems apply pruning after model training, where a significant amount of computation is required to fine-tune the remaining parameters to regain accuracy~\cite{pavlo2017}. \Reconv preserves model accuracy by retaining important parameters by applying unstructured pruning at model initialization. In addition, less significant layers of a DNN, which contain parameters that least contribute to accuracy, undergo structured pruning. \Reconv adopts a disciplined approach to determine the significance of parameters and their contribution to the importance of DNN layers. This allows for a more precise structured pruning approach that reduces model size and accelerates training and inference while maintaining accuracy. 

\begin{table}[t]
\caption{Comparing unstructured/structured pruning (UP, SP), neural architecture search (NAS), and \Reconvv.}
	\label{tab:comp}
    \centering
    \begin{tabular}{p{4cm}cccc}

\hline
                  \scriptsize{Characteristics of compressed models} & \footnotesize{UP} & \footnotesize{SP} & \footnotesize{NAS} & \footnotesize{\Reconvv} \\ \hline
\footnotesize{Maintain high accuracy}    &  \cmark   &  \xmark    &     \cmark      &     \cmark       \\
\footnotesize{Smaller and faster}     &  \xmark  &  \cmark  &    \cmark      &     \cmark       \\
\footnotesize{Rapidly discovered} &  \cmark   &  \cmark  &  \xmark  &  \cmark               \\\hline

\end{tabular}
\vspace{-10pt}
\end{table}  

\Reconv produces pruned models with seconds at model initialization that are up to 16.21$\times$ smaller and 2$\times$ faster while maintaining the same accuracy to the dense model. 
Our \textbf{research contributions} are as follows:
\begin{itemize}
    \item The development of \Reconvv, a system that is underpinned by a novel pruning at initialization (PaI) method for convolutional neural networks to determine which layers are sensitive to structured pruning systematically.
    \item The experimental demonstration that selective structured pruning based on layer sensitivity maintains accuracy on par with unstructured PaI methods. 
    \item The empirical demonstration that structured PaI can be used to rapidly search for optimized pruned models and then train edge DNNs with lower resource overheads than neural architecture search.
\end{itemize}

The remainder of this paper is organized as follows. Section~\ref{sec:background} discusses background content to model pruning. Section~\ref{sec:reconvene} presents the \Reconv system. Section~\ref{sec:studies} presents the experimental results. Section~\ref{sec:relatedwork} discusses related work, and Section~\ref{sec:conlusion} concludes the paper.

\section{Background}
\label{sec:background}
\begin{figure*}[!t]
\centerline{\includegraphics[width=0.77\textwidth]{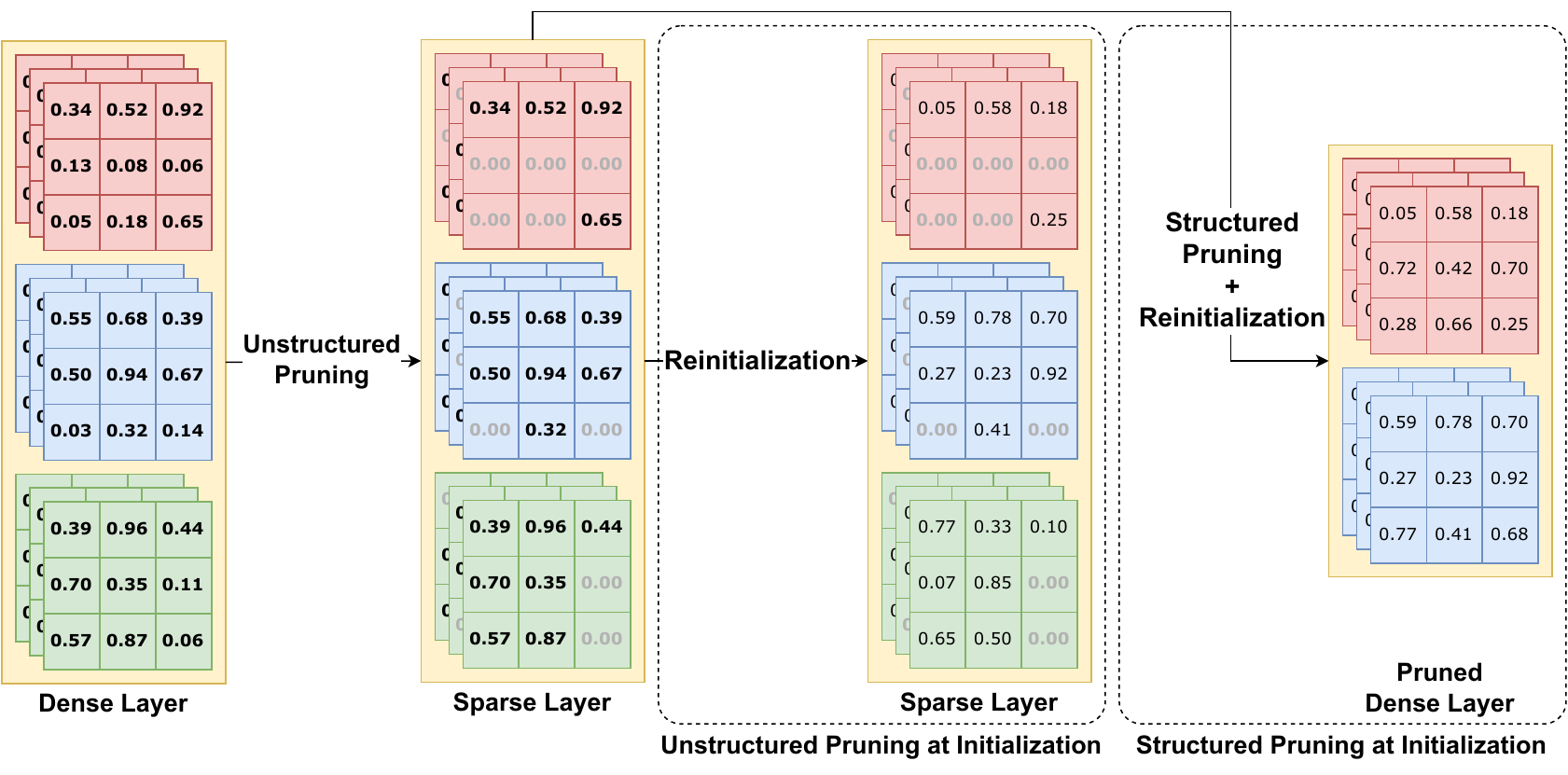}}
\caption{Different pruning at initialization (PaI) methods applied to a convolutional layer. UPaI prunes and then reinitializes the remaining parameters. SPaI redistributes the parameters such that a smaller layer of only dense channels is created.}
\label{fig2}
\vspace{-8pt}
\end{figure*}

In this section, we present pruning methods, specifically unstructured and structured pruning, in addition to pruning at initialization (PaI) and its relevance to neural architecture search (NAS) when searching for lightweight models within a large DNN suitable for deployment at the edge.

\subsection{Unstructured and Structured Pruning}
DNN pruning aims to reduce the computational complexity of models by removing redundant parameters (weights or connections). An ideal pruning method will prioritize the removal of parameters that contribute the least to model accuracy for maintaining usability after compression. Pruning methods are categorized as \textit{Unstructured} and \textit{Structured} pruning.

\textbf{Unstructured Pruning}
(UP) masks individual parameters by setting their value to zero~\cite{NIPS1989_6c9882bb, han2015}. A ranking algorithm determines which parameters to mask using simple metrics such as the magnitude of the weights~\cite{lth}, to more complex criteria using training information~\cite{synflow, prospr}. By masking parameters, the model becomes sparse, referred to as a sparse model, and the original model is referred to as a dense model. While UP maintains model accuracy between $\sim$50-90\% sparsity depending on the model, dataset, and pruning method~\cite{pruning_survey}, sparse models only provide runtime performance improvements in cloud scenarios~\cite{gale2019state} or where libraries for sparse matrix formats are available~\cite{sparselib2, unisparse}.

Edge devices may not be equipped with hardware accelerators, such as GPUs~\cite{edge_survey}, or may not support sparse matrix representations and libraries~\cite{sparselib2, prunefl}. Consequently, sparse models on the edge have limitations. Firstly, scattered sparsity in dense convolutions leads to irregular memory access patterns, which hinder both model training and inference~\cite{benefits}. Secondly, since zeroed parameters still consume the same memory as non-zero parameters, there is no gain in memory efficiency~\cite{ECCLES2023, sparselib2}. Figure~\ref{fig1} summarizes UP: high accuracy at the cost of little performance benefits.

\textbf{Structured Pruning}
(SP) removes groups of parameters such as filters, channels, or layers~\cite{wen2016learning, he2017}. SP results in a spatially smaller pruned model~\cite{li2017pruning}, beneficial to edge scenarios with a high demand for models with low memory, energy, and inference footprints~\cite{pavlo2017}. However, obtaining high-quality pruned models is challenging since: (1) SP is oriented towards runtime performance improvements. Therefore, profiling every prospective model from a large search space can take hours~\cite{easiedge} to days~\cite{pavlo2017} to find a single high-quality pruned model. (2) At higher sparsities, essential parameters are inevitably removed; fine-tuning is required to regain accuracy, which can take many times longer than the original model training time for complex datasets~\cite{han2015}; instead, training a new model of the same size from scratch may result in better accuracies~\cite{liu2018rethinking}. Figure~\ref{fig1} summarizes SP: Improved runtime performance at the cost of model accuracy.

\subsection{Pruning at Initialization}
\label{bg:pai}
Typically pruning occurs after~\cite{NIPS1989_6c9882bb, he2017} or during~\cite{gale2019state} model training. However, recent pruning literature explores pruning at network initialization (PaI), where before training, it is possible to discover a sub-network of randomly initialized parameters that, when fully trained, can match the accuracy of the original dense network~\cite{lth}. Existing PaI literature focuses on unstructured PaI. However, recently, the feasibility of structured PaI has been explored~\cite{prospr, cai2022structured}.

\textbf{Unstructured Pruning at Initialization} (UPaI) involves the UP of a dense network, then reinitializing the remaining parameters before training~\cite{lth}. UPaI can match the accuracy within 1\% of a dense model up to $\sim$98\% sparsity~\cite{synflow, frankle2021pruning, Wang2020Picking}. The first three stages in Figure~\ref{fig2} show the generalized approach of UPaI. While UPaI presents the opportunity to accelerate training using the sparse model as a drop-in replacement to the original dense model, it encounters challenges in edge scenarios for the same reasons as UP~\cite{ECCLES2023}.

\textbf{Structured Pruning at Initialization} (SPaI) extends UPaI for improved runtime performance. While UPaI produces a sparse model, SPaI introduces an additional step before reinitialization: the model is pruned using SP. First, this SP spatially compresses the model, and second, sparse layers are converted into dense layers of the same parameter count, which improves hardware utilization~\cite{repvgg}. The first two and the last column in Figure~\ref{fig2} show SPaI. This example converts a 33\% sparse 3-channel layer into a dense 2-channel layer with the same parameter count. SPaI presents an opportunity for edge-compatible pruned models to be discovered within seconds, significantly outperforming NAS in search time~\cite{ofa}. In addition, SPaI has considerably lower overheads than NAS~\cite{cai2022structured}, allowing for execution on an edge device to create pruned models tailored for the device~\cite{perf_predict, ECCLES2023}.

\begin{figure}[!t]
\centerline{\includegraphics[width=0.49\textwidth]{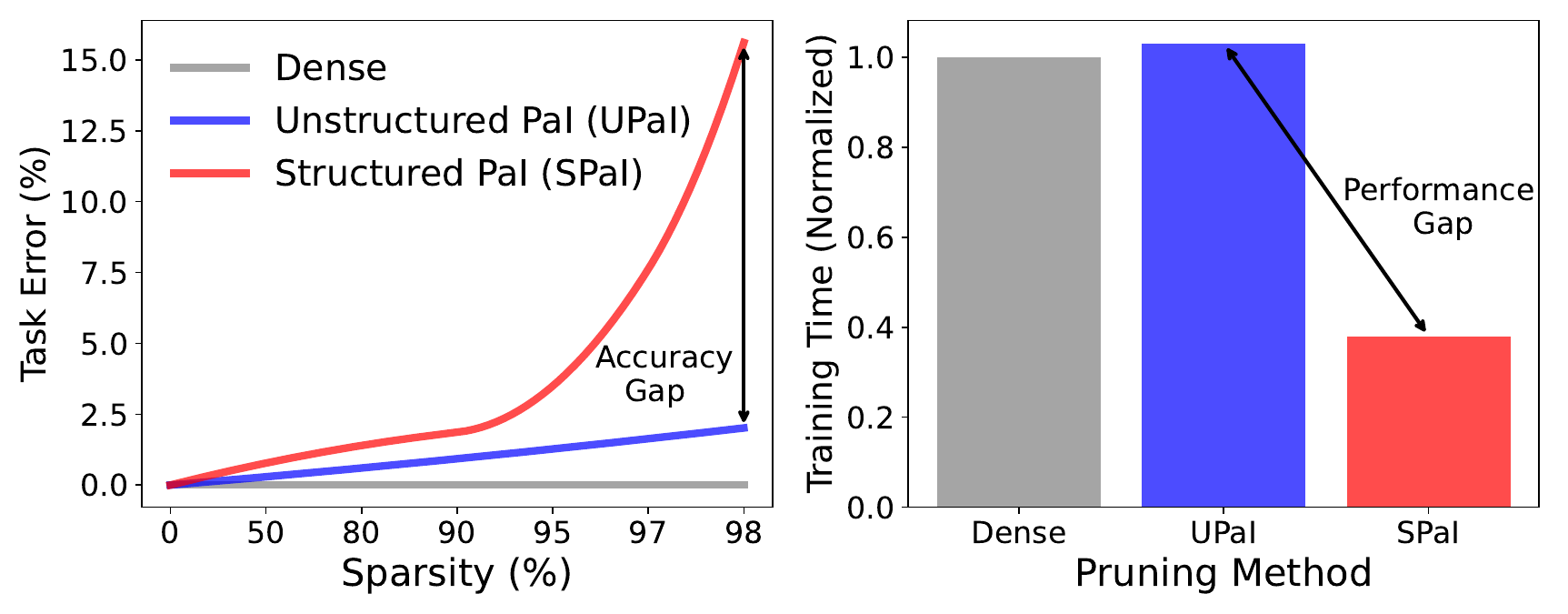}}
\caption{Pruning at initialization (PaI) methods for VGG-16 (CIFAR-10). UPaI maintains model accuracy without improving runtime performance, while SPaI improves performance but reduces accuracy.}
\label{fig3}
\vspace{-10pt}
\end{figure}

\begin{figure*}[!t]
\centerline{\includegraphics[width=0.9\textwidth]{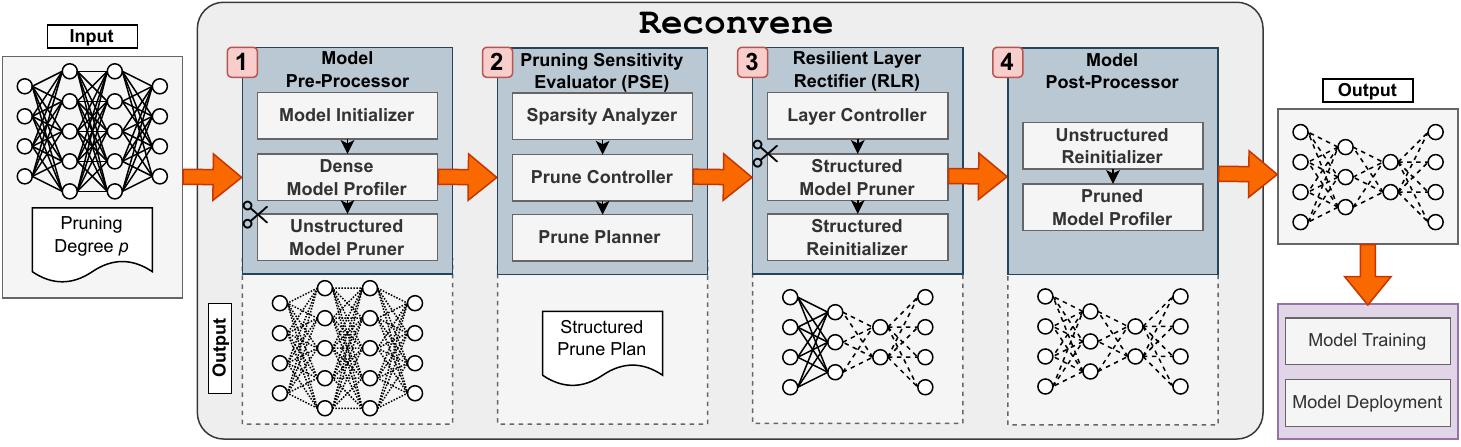}}
\caption{System overview of \Reconvv.}
\label{fig4}
\vspace{-10pt}
\end{figure*}

\textbf{Structured PaI Challenges:} When implementing SPaI as described in Figure~\ref{fig2} raises the following questions: \textit{Do dense layers from SPaI achieve the same accuracy as sparse layers from UPaI when both have the same number of parameters?} Recent literature suggests that where an individual parameter is located within a layer holds no significance for UPaI; instead, the layer-wise sparsity ratio is more critical to model accuracy~\cite{frankle2021pruning, cai2022structured}. Therefore, SPaI, in theory, should achieve close to, or the same, accuracy as UPaI. Figure~\ref{fig3} shows that SPaI maintains accuracy close to UPaI up to $\sim$90\% before quickly collapsing. This generally holds true for many models and datasets~\cite{frankle2021pruning}. However, achieving higher sparsity SPaI ($>$90\%) with matching accuracy to UPaI will allow for smaller pruned models to be discovered (within seconds) and trained (in a fraction of the time of dense) on edge devices. Currently, insights are limited on SPaI~\cite{cai2022structured}. \textit{This paper focuses on addressing this challenge - minimizing the accuracy and performance gap between UPaI and SPaI.}

\section{Reconvene}
\label{sec:reconvene}
\Reconv is the proposed system to facilitate rapid pruning of DNN models for edge deployment. It is underpinned by an SPaI method that closely matches UPaI accuracies. This produces highly compressed models with low training/inference latencies. \Reconv can be used to deploy a model across a range of heterogeneous edge devices where each model instance is selectively pruned on initialization for that device. \Reconv is suitable for use cases that require models to be tailored to resource availability or capability for federated learning using small devices~\cite{prunefl, zerofl, gradfl}. 

\subsection{Motivation}
Existing model pruning systems generally comprise only one of the two types of pruning. In cloud-based systems, the emphasis is on model accuracy since significant computational resources are available~\cite{lth}. However, the efficiency of model discovery and model latency must be considered given the operational costs and constraints~\cite{ofa}. On the other hand, when it comes to edge computing, model pruning methods aim to balance model accuracy with the stringent resource constraints of devices. While accuracy is reduced, the goal is typically to achieve significant model compression without compromising performance~\cite{han2015}. In recent years, hybrid systems that utilize both unstructured and structured pruning have been shown to strike a balance between accuracy and computational demands post-training~\cite{ECCLES2023, subfedavg}. SPaI aims to achieve a balance similar to post-training hybrid pruning methods. However, PaI also offers the advantage of improved training efficiency~\cite{cai2022structured, prospr}. Thus, SPaI enables models to be trained on edge devices that have limited computational and memory resources. In cloud environments, it can reduce operational costs~\cite{ofa}.

Existing SPaI methods~\cite{cai2022structured} have the following limitations that significantly reduce model accuracy. First, they fully reparameterize sparse models into pruned models, thereby removing the fine-grained accuracy-preserving properties of unstructured pruning~\cite{frankle2021pruning, cai2022structured}. Second, they apply the same pruning method to all model layers. This inherently prunes important layers while under-pruning redundant layers~\cite{frankle2021pruning, cai2022structured}. These limitations have led to SPaI models with worse accuracy than training a smaller model from scratch~\cite{cai2022structured}.

\Reconv addresses the above limitations by incorporating a two-step process to SPaI that determines how sensitive each layer is to structured pruning. This allows for \Reconv to control the amount and type of pruning of each layer to maximize model compression while minimizing accuracy loss.

\subsection{System Overview}
Figure~\ref{fig4} provides an overview of \Reconv comprising four modules: \textbf{\textit{Model Pre-Processor}}, \textbf{\textit{Pruning Sensitivity Evaluator} (PSE)}, \textbf{\textit{Resilient Layer Rectifier} (RLR)}, and \textbf{\textit{Model Post-Processor}} as part of the SPaI pipeline. The PSE and RLR modules prepare a pruned model for edge training and deployment, especially considering that SPaI, on its own, does not maintain model accuracy with increased levels of pruning (Figure~\ref{fig3}). Each module is detailed below.

\textbf{Input:} The end-user chooses a dense model and a pruning degree $p \in [0,1]$. For example, $p=0.8$ prunes 80\% of the model parameters.

\textbf{Model Pre-Processor:} The input model undergoes UP to the pruning degree $p$ in this module using the following three components. First, the model is initialized in memory as a dense model via the \textit{Model Initializer} component that loads the dense input weights into the chosen model architecture. Next, the \textit{Dense Model Profiler} is used to gather runtime metrics of the original model. Using a synthetic image input over several samples, the dense model is profiled for memory consumption, model size, and CPU/GPU latency. Finally, the \textit{Unstructured Model Pruner} utilizes magnitude pruning~\cite{lth} to apply unstructured pruning to the dense model. The output from this component (and module) is a sparse model pruned to the degree $p$. \Reconv is designed to be interoperable with existing model pruning systems. As such, the unstructured pruning method is fully configurable. Magnitude pruning is used by default as it is effective across a wide range of model architectures and datasets~\cite{frankle2021pruning}. However, this can be substituted for any other UP method, such as SynFlow~\cite{synflow}, by user choice.

\begin{figure}[!t]
\centerline{\includegraphics[width=0.45\textwidth]{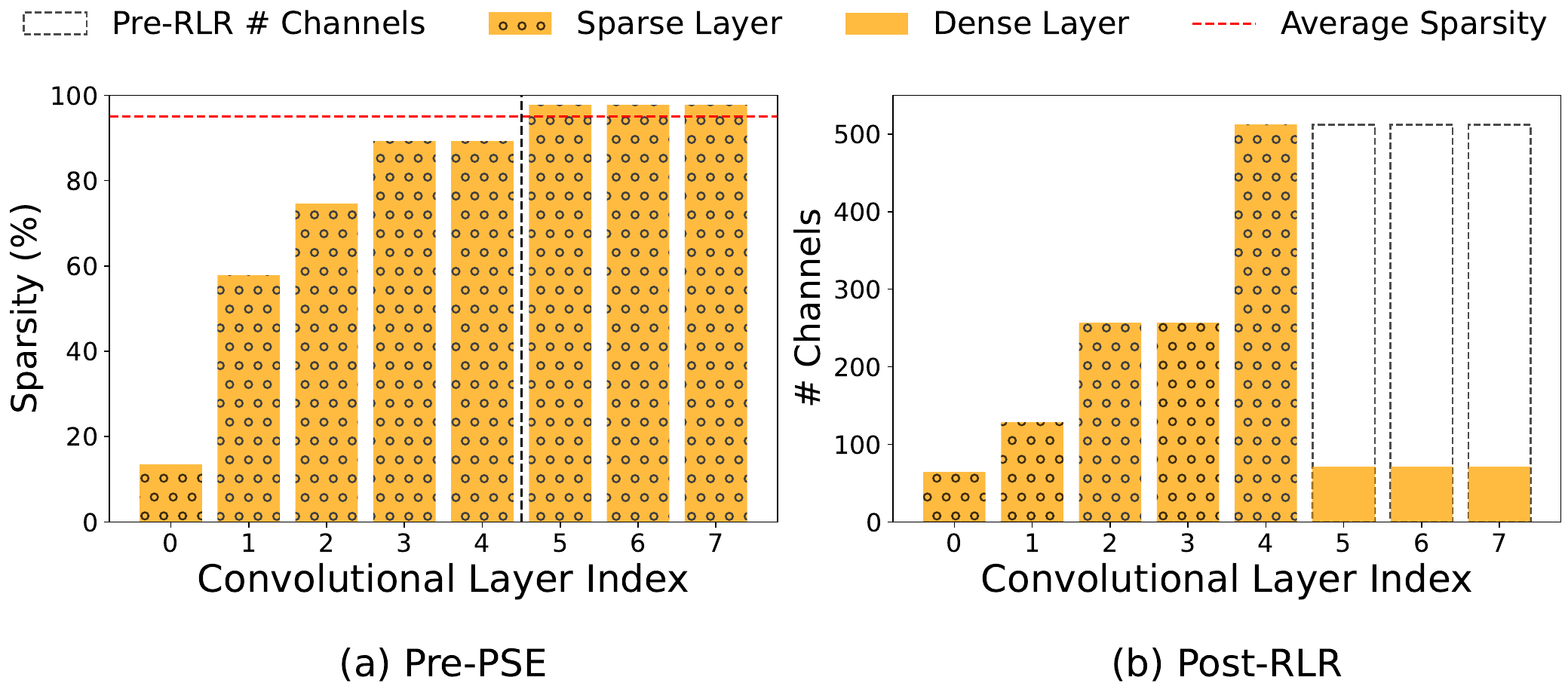}}
\caption{VGG-11 (CIFAR-10, $p=0.95$) before and after \Reconvv. Layers to the left of the vertical dashed line are considered to be sensitive to pruning and, thereby, do not undergo structured pruning in the RLR.}
\label{fig5}
\vspace{-10pt}
\end{figure}

\textbf{Pruning Sensitivity Evaluator (PSE):} PSE evaluates the sensitivity of each layer to pruning by contrasting its sparsity with the average sparsity of the global model. First, the \textit{Sparsity Analyzer} calculates the per-layer sparsity of the model by examining the sparsity pattern of the sparse model created by the \textit{Unstructured Model Pruner}. Next, the \textit{Prune Controller} determines the sensitivity of each layer to pruning by comparing the sparsity of that specific layer with the average sparsity of the sparse model. Layers with minimal unstructured pruning (low sparsity, under the average) are regarded as sensitive to pruning, while resilient layers contain a large number of redundant parameters and are unaffected by unstructured pruning. Figure~\ref{fig5} (a) shows the state of various convolution layers after unstructured pruning but before PSE, named pre-PSE. Each layer has a different sparsity, and the red line is the average sparsity of the model. In the \textit{Prune Controller}, sensitive layers are those which have a lower sparsity than the model average, whereas resilient layers are those which have a higher sparsity than the average. Determining layer sensitivity is captured in the following equation where $S_l$ is the sparsity of layer $l$ and $S_{\text{Avg}}$ is the average model sparsity:
\begin{equation} \label{eq:1}
    PSE(l) \begin{cases} 
      False & S_l \geq S_{\text{Avg}} \\
     True & S_l < S_{\text{Avg}}
   \end{cases}
\end{equation}
Finally, the \textit{Prune Planner} calculates how much each resilient layer should be pruned via structured pruning as the product of the number of channels in the layer by the sparsity of the layer (Section~\ref{subsec:impl})  and passes this prune plan to the RLR.

\textbf{Resilient Layer Rectifier (RLR):} The RLR applies structure pruning to the resilient layers. Using the structured pruning plan from PSE and \textit{Layer Controller}, RLR applies SPaI to only the resilient layers. First, the number of channels is reduced proportionally to the layer density using the \textit{Structured Model Pruner}. Next, the remaining channels are reinitialized as dense parameters using the \textit{Structured Reinitializer}. For instance, a sparse layer with 100 channels, exhibiting a 95\% sparsity following UP, is transformed into a dense layer consisting of 5 channels. Figure~\ref{fig5} (b) illustrates the post-RLR state of the convolutional layers in Figure~\ref{fig5} (a). The extent to which each resilient layer is pruned following SPaI is illustrated by the dashed outline, which indicates the number of channels present before the application of SPaI.

\textbf{Model Post-Processor:} Post-RLR, the remaining sensitive layers are reinitialized so that the model is completely reinitialized with pseudo-random Kaiming weights~\cite{frankle2021pruning} to complete pruning at initialization using the \textit{Unstructured Reinitializer}. After which, the remaining model contains a mix of reinitialized sparse and pruned layers and is ready for training and deployment. In addition, the pruned model is profiled for the metrics seen in the pre-processor to gather empirical speedup and compression metadata. The metadata determines what category of edge device the pruned model can be trained and deployed to.

\textbf{Output:} \Reconv outputs an initialized model pruned by $p$, which is ready for training and edge deployment using typical training hyperparameters (see Table~\ref{tab:hparams}).

\begin{algorithm}[t]
\DontPrintSemicolon
\caption{Pruning Sensitivity Evaluator (PSE)}
\label{alg1}

\KwData{Convolutional layer $l$ with a weights matrix $W$ with shape $(N_{\text{channels}}, N_{\text{weights}})$, Global average sparsity of the model $S_{\text{Avg}}$}
\KwResult{Number of channels for layer $l$}

$N_{\text{total}} \gets 0$; $N_{\text{zero}} \gets 0$; $i \gets 0$\;
\While{$i < N_{\text{channels}}$}{
    $j \gets 0$\;
    \While{$j < N_{\text{weights}}$}{
        $N_{\text{total}} \gets N_{\text{total}} + 1$\;
        \If{$W_{i,j} == 0$}{
            $N_{\text{zero}} \gets N_{\text{zero}} + 1$\;
        }
        $j \gets j + 1$\;
    }
    $i \gets i + 1$\;
}
$S_l \gets \frac{N_{\text{zero}}}{N_{\text{total}}}$\ \tcp*[r]{Pruning degree of $l$}
\If{$S_l \geq S_{\text{Avg}}$}{
    \Return{} $\lceil N_{\text{channels}} \cdot S_l \rceil$ \tcp*[r]{Resilient layer}
} \Else {
    \Return{} $N_{\text{channels}}$ \tcp*[r]{Sensitive layer}
}

\end{algorithm}

\begin{algorithm}[t]
\DontPrintSemicolon
\caption{Resilient Layer Rectifier (RLR)}
\label{alg2}
\KwData{Original unstructured sparse model \( M \), List of resilient layer indices \( R \), List of pruned channel sizes for resilient layer indices \( C \)}
\KwResult{Reinitialized pruned model \( M^\prime \)}
\(M^\prime \gets M\) \tcp*[r]{Initialize \( M^\prime \) based on \( M \)}

\ForEach{layer \( l \) in \( M \)}{
    \If{\( l \) is in \( R \)}{
        \( M^\prime_l \) $\gets$ Structured Prune(\( l, C_l \))
        
        \( M^\prime_l \) $\gets$ Reinitialize(\( M^\prime_l \))
    }
}
\Return{} \( M^\prime \)

\end{algorithm}

\subsection{Implementation}
\label{subsec:impl}
\Reconv is implemented using Python 3.11.4, PyTorch~2.0.1, Torchvision 0.15.2, and CUDA 11.7 and is intended to be used as a straightforward and lightweight system for determining the amount to which each layer needs to be pruned during SPaI with the objective of edge-centric pruned models. The \Reconv PSE and RLR algorithms are provided below as Algorithm~\ref{alg1} and Algorithm~\ref{alg2}, respectively.

\textbf{Pruning Sensitivity Evaluator (PSE):} Algorithm~\ref{alg1}, is the first step in preparing a sparse model $M$ for SPaI. PSE has two sub-steps: Step 1 - Calculate the layer-wise sparsity of each layer, and Step 2 - Adjust the number of channels in each SPaI layer. Step 1 is achieved by iterating through the unstructured pruned layers and totaling the number of non-zero and zero parameters (sparse parameters) - Algorithm~\ref{alg1} Line 1-12. The sparsity of the layer, or $S_l$, is the fraction of zero parameters divided by total parameters - Algorithm~\ref{alg1} Line 13. At this point, layer $l$ only contains unstructured sparsity. Nonetheless, through SPaI, the objective is to create a structured pruned layer, denoted $l^\prime$, while maintaining the same parameter count as $l$. This is achieved by reducing the number of channels, $N_{\text{channels}}$ in $l$ proportionally to $S_l$. To accomplish this, PSE initially assesses whether $l$ should be pruned by computing its sensitivity according to Equation~\ref{eq:1}, as outlined in Algorithm~\ref{alg1} Line 14. If $l$ is resilient, then $N_{\text{channels}}$ is scaled by $S_l$ and rounded up to the nearest full channel - Algorithm~\ref{alg1} Line 15. Otherwise, the layer is sensitive and remains untouched - Algorithm~\ref{alg1} Line 18.

\textbf{Resilient Layer Rectifier (RLR):} Algorithm~\ref{alg2}, is the second step in \Reconv by applying SPaI and reinitialization of the remaining parameters. A copy of which layers are resilient to pruning and the number of channels ($N_{\text{channels}}$) they should be reduced to are provided as lists $R$ and $C$, respectively. Next, sparse model $M$ is iterated, and for each resilient layer, $l$ in $R$ undergoes structured pruning to the size $C_l$ - Algorithm~\ref{alg2} Line 4. Finally, the pruned layer is fully reinitialized with random dense parameters - Algorithm~\ref{alg2} Line 5. Each step updates the pruned model reference $M^\prime$ and is then returned - Algorithm~\ref{alg2} Line 8.

\section{Experiments}
\label{sec:studies}
This section first presents the experimental setup in Section~\ref{exp:setup} and then considers four aspects of \Reconvv:

(1) \textbf{Method Validation}: Validating \Reconv modules, namely \textit{Pruning Sensitivity Evaluator (PSE)} and \textit{Resilient Layer Rectifier (RLR)} across a range of sparsities, models, and datasets for evaluating model accuracy, compression, and CPU/GPU speedup (Section~\ref{exp:ablation}).

(2) \textbf{Training Benefits}: Evaluating the accelerated training time and final accuracy of pruned models created with \Reconv and comparing them against existing SPaI/UPaI methods and training a smaller model from scratch (Section~\ref{exp:training}).

(3) \textbf{Pruned Model Quality}: Comparing \Reconv to existing SPaI pruning systems and the runtime performance metrics of the pruned models across a range of sparsities (Section~\ref{exp:systems}).

(4) \textbf{System Overheads}: Contrasting the low overheads of \Reconv against exhaustive NAS methods examining both total time and memory requirements for creating compressed models (Section~\ref{exp:overheads}).

\begin{table}[t]
\caption{Baseline model results and training hyperparameters for a production quality dense VGG-16, ResNet-20, and ResNet-50 models.}
\label{tab:hparams}
\centering
\begin{tabular}{lrrr}
\hline
                & \multicolumn{1}{r}{VGG-16} & \multicolumn{1}{r}{ResNet-20} & \multicolumn{1}{r}{ResNet-50}     \\ \hline
Dataset         & CIFAR-10  &  CIFAR-10               & Tiny ImageNet \\
Parameters (M)     & 14.72      &    0.27           & 25.56                             \\
Size (MB)       & 56.2        &      1.1         & 100.1                              \\
Accuracy (\%) & 93.32        &    91.68          & 55.48                               \\
\# Epochs        & 160             &    160       & 200                               \\
Batch Size     & 128             &   128        & 256                               \\
Learning Rate   & 0.1             &  0.1        & 0.2                               \\
Milestone Steps\tablefootnote{Learning rate drops by a factor of gamma, 0.1, at each milestone step. All models use the SGD optimizer, momentum of 0.9 and weight decay of 0.0001.} & 80, 120       &       80, 120      & 100, 150                         \\ \hline
\end{tabular}
\vspace{-8pt}
\end{table}

\subsection{Experimental Setup}
\label{exp:setup}
Three DNN models trained on the CIFAR-10~\cite{cifar10} and Tiny ImageNet~\cite{le2015tiny} datasets are considered. The first is VGG-16~\cite{vgg} trained on CIFAR-10, serving as a straightforward feedforward model. The other two, ResNet-20 and ResNet-50~\cite{resnet}, are trained on CIFAR-10 and Tiny ImageNet, respectively, representing models with branching structures. These models and datasets are widely recognized for their production quality and are frequently employed as benchmarks in pruning literature\cite{pruning_survey}. Some experiments utilize VGG-19 (including \Reconvv) when other methods do not report VGG-16 as a baseline in the literature. In addition, alternate versions of each baseline are used to compare pruned large models to smaller dense versions. For example, VGG-11 and ResNet-8 are a shallower version of VGG-16~\cite{vgg} and ResNet-20~\cite{resnet}, respectively.

\textbf{Models, Datasets, and Hyperparameters:} The VGG models are those that have one fully connected layer~\cite{lth}, and ResNets are the default configurations~\cite{resnet} for their respective datasets. CIFAR-10 consists of 50,000 training images and 10,000 test images of the dimension 32$\times$32$\times$3 divided equally across 10 classes. Tiny ImageNet is a subset of ImageNet consisting of 100,000 training images and 10,000 test images of the dimension 64$\times$64$\times$3 divided equally across 200 classes. The baseline results are obtained using the training routine of OpenLTH (a PaI framework)~\cite{lth} using the hyperparameters in Table~\ref{tab:hparams}.

\textbf{Testbed:} We use an AMD EPYC 7713P 64-core/128-thread CPU and two Nvidia RTX A6000 GPUs to train, profile, and prune the Tiny ImageNet models, as such resources are representative of those in a cloud data center. CIFAR-10 experiments are carried out with an Intel i9-13900KS 24-core/32-thread CPU and an Nvidia RTX 3080 GPU comparable to an edge server that may be used in a production setting.

\textbf{Trial Counts and Reporting Methods:} All experiments were carried out three times. Model performance metrics, such as accuracy, memory usage, and latency, are presented in tables and figures as the mean from all experiments accompanied by confidence intervals spanning one standard deviation.

\textbf{Pruning Setup:} Pruning experiments are carried out across six different sparsities \{50, 80, 90, 95, 97, 98\} grouped into three difficulties from easy to hard. Trivial sparsities (Easy) \{50, 80\} are those in which even random pruning will match unstructured pruning. Matching sparsities (Medium) \{90, 95\} are those in which benchmark methods perform well and can still match the unpruned dense model accuracy. Extreme sparsities (Hard) \{97, 98\} are those in which the accuracy of the models generated by unstructured pruning methods is lower than unpruned dense models.

\begin{figure*}[t!]
\centerline{\includegraphics[width=0.98\textwidth]{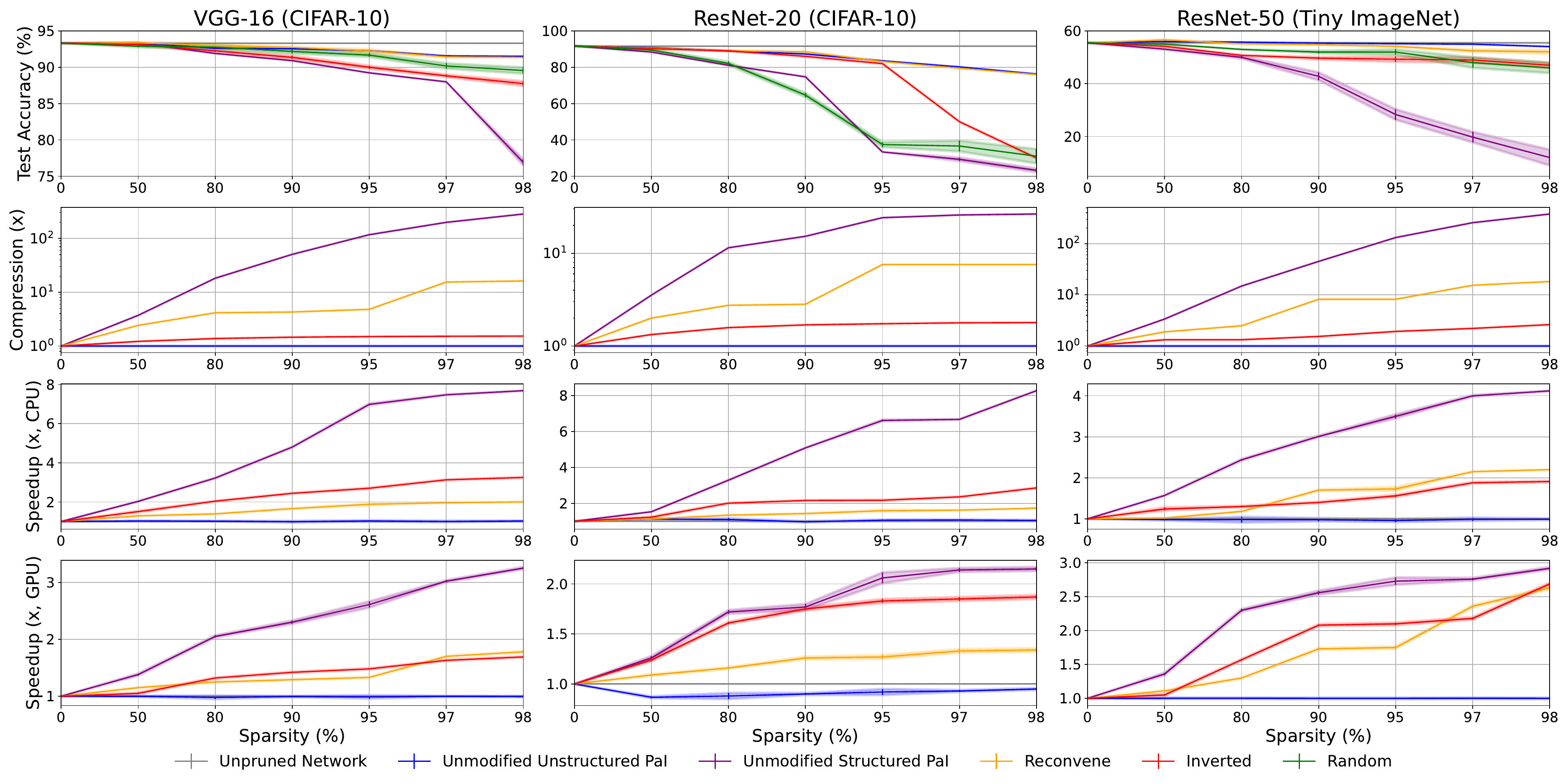}}
\caption{Comparing test accuracy, model compression, and CPU/GPU speedup across a range of sparsities for a selection of UPaI and SPaI pruning methods for VGG-16 and ResNet-20 models on CIFAR-10 and ResNet-50 on Tiny ImageNet datasets. Random is plotted only for accuracy tests because of its large variance in compression and speedup during random pruning.}
\label{fig6}
\vspace{-8pt}
\end{figure*}

\textbf{Pruning Metrics:} The metrics reported in this paper are similar to those adopted in the literature~\cite{pruning_survey, gale2019state, wang2023state, fang2023state}. Accuracy is reported as absolute top-1 test accuracy when comparing pruned models from the same baseline model or relative top-1 test accuracy \textit{delta} ($\Delta$) when the baseline model accuracy can not be replicated across different pruning systems~\cite{pruning_survey, fang2023state}. The mean model inference speedup of three trials is calculated as $\frac{\text{Mean Dense Model Latency}}{\text{Mean Compressed Model Latency}}$. Compression is calculated as $\frac{\text{Dense Model Size}}{\text{Compressed Model Size}}$ where model size is the size of the model in storage. Note that this is different from using FLOPs and parameter count to calculate theoretical speedup and model size, respectively~\cite{pruning_survey}. We have utilized performance metrics that are empirically observed to showcase the practical benefits in real-world deployment scenarios~\cite{deployment}.

\subsection{Validation of \Reconvv}
\label{exp:ablation}
In this experiment, the PSE module for determining layer sensitivity (Equation~\ref{eq:1}) is empirically validated against baseline methods considered below, as well as an inverted sensitivity method to demonstrate that \Reconv maintains the same model accuracy as UPaI methods across a range of sparsities. Thereby, all confounding variables are eliminated, such as differences in training hyperparameters~\cite{wang2023state} or model architecture~\cite{pruning_survey}. Each method is tested on the same baseline model, testbed, and set of hyperparameters. The methods considered in Figure~\ref{fig6} are as follows: (1)~\textit{Unpruned Network:} Dense fully trained model representing maximum model accuracy, and is used as the reference for compression and speedup calculations. (2) \textit{Unmodified UPaI:} Sparse model pruned with a magnitude-based unstructured pruning method~\cite{lth} as described in Section~\ref{bg:pai}. (3) \textit{Unmodified SPaI:} Pruned model following reparameterization of the \textit{Unmodified UPaI} model as described in Figure~\ref{fig2}. (4) \textit{Reconvene:} \Reconv as described in Section~\ref{sec:reconvene}. (5) \textit{Inverted:} \Reconvv, however, the underlying method of the PSE Prune Controller component is inverted. In other words, structured pruning occurs on sensitive layers in the RLR. (6) \textit{Random:} \Reconvv, however, the PSE Prune Controller randomly assigns layers as sensitive.

Figure~\ref{fig6} shows the results for each method across multiple model architectures and datasets. For VGG-16 (CIFAR-10), for all levels of sparsity, \Reconv is on par with UPaI for accuracy. The SPaI, inverted, and random methods, however, see a decline in accuracy beyond trivial sparsity levels of 80\%. At 98\% sparsity, the accuracy of the random method is 2\% less than both \Reconv and UPaI, while SPaI and inverted lag behind by 15\% and 4\%, respectively. This data supports the observation that \Reconv is selectively pruning the right layers to maintain accuracy. In contrast, the inverted method, which prunes the opposite set of layers, reduces accuracy. Additionally, employing unmodified SPaI to prune all layers results in the highest decline in model accuracy.

\begin{figure*}[t!]
\centerline{\includegraphics[width=0.98\textwidth]{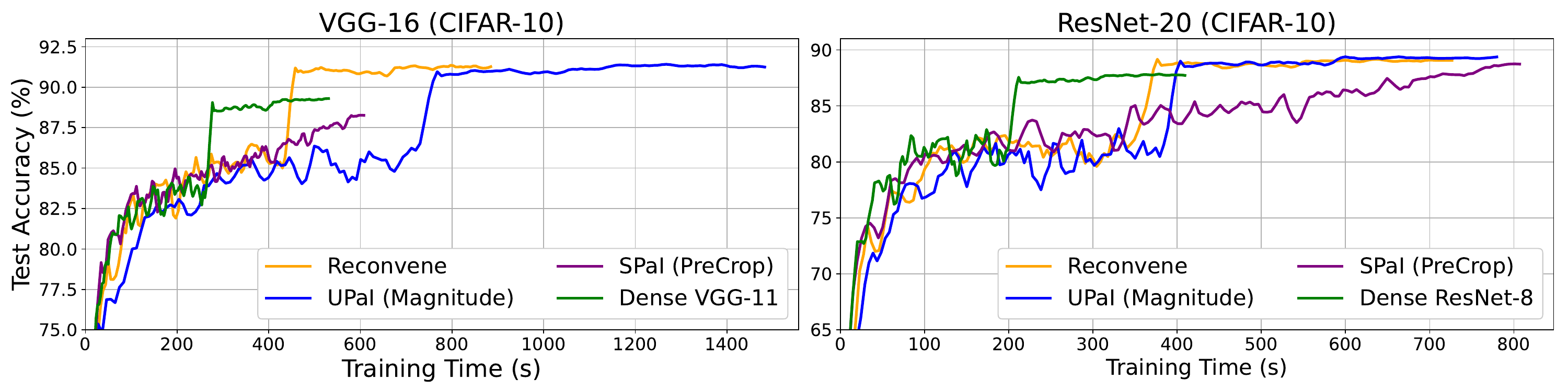}}
\caption{Comparison of test accuracy over training time for VGG-16 (CIFAR-10, $p=0.98$) and ResNet-20 (CIFAR-10, $p=0.8$) for various UPaI and SPaI pruning methods and alternative smaller models (VGG-11 and ResNet-8).}
\label{fig7}
\vspace{-8pt}
\end{figure*}

In terms of model compression, UPaI remains at 1$\times$ for all sparsity levels since zero parameters are the same size in storage as non-zero parameters. SPaI obtains the highest compression, but the model becomes impractical to use at extreme sparsity levels because of its diminished accuracy. \Reconv achieves the second highest compression, primarily because resilient layers are typically found deeper in the DNN, and these layers are usually larger than the initial ones. On the other hand, the inverted method targets the more sensitive layers, which are generally smaller and positioned at the beginning of the DNN. Consequently, this results in a compression rate lower than \Reconvv.

For CPU/GPU speedup, a similar trend is observed for SPaI, achieving the highest speedup but at the cost of model usability. Since the early layers, which the inverted method prunes, are typically slower than later layers, the inverted method obtains a higher speedup than \Reconvv. However, this is only observed in the CPU inference tests. The GPU inference test shows that \Reconv and the inverted method obtain nearly the same speedup at most sparsity levels. For UPaI, without specialized hardware accelerators, sparse models cannot be effectively utilized. When accelerators are not available, a sparse model can sometimes be slower than its dense counterpart (up to 2\% slower; Figure~\ref{fig6}).

The same trend is observed for both ResNet models on CIFAR-10 and Tiny ImageNet. \Reconv maintains the same accuracy as UPaI for ResNet-20 with a small divergence at extreme sparsities for ResNet-50. In addition, \Reconv reaches 8$\times$ and 18.2$\times$ compression for ResNet-20 and ResNet-50 at 98\% sparsity with $\sim$2$\times$ CPU speedup on both models and 1.34$\times$ and 2.63$\times$ GPU speedup respectively.

\textbf{Observation 1}: Without sacrificing the model accuracy achieved by UPaI, \Reconv obtains up to 16.21$\times$ compression, 2$\times$ and 1.78$\times$ CPU and GPU speedup, respectively.

\subsection{Improvements during training}
\label{exp:training}
In this experiment, \Reconv is compared against a SPaI method, namely PreCrop~\cite{cai2022structured}, UPaI, and training a smaller model of the same architecture from scratch. The aim is to demonstrate that: (1) \Reconv trains a model to full accuracy faster than UPaI, (2) \Reconv trains to a higher accuracy than SPaI, and (3) \Reconv creates pruned models which train faster and to a higher accuracy than manually choosing a smaller model and training it from scratch.

Figure~\ref{fig7} shows the results for VGG-16 and ResNet-20. For VGG-16, UPaI takes 1,482 seconds to train to an accuracy of 91.24\%. \Reconv takes 884 seconds, 1.68$\times$ faster than UPaI, to train to an accuracy of 91.26\%. PreCrop, which applies SPaI to all layers, trains to an accuracy of 88.26\% in 607 seconds. Within the same timeframe, \Reconv has already trained to an accuracy of 90.6\%. Furthermore, PreCrop trains more slowly and achieves a lower accuracy compared to VGG-11 at 89.3\% after 531 seconds. In this scenario, choosing VGG-11 for deployment instead of pruning VGG-16 with PreCrop would yield a higher-quality model in a shorter time. Conversely, \Reconv achieved 91.1\% accuracy in just 440 seconds of training, outperforming both VGG-11 and PreCrop.
Similarly, for ResNet-20, UPaI takes 780 seconds to train to an accuracy of 89.37\%. \Reconv takes 725 seconds, 1.08$\times$ faster than UPaI, to train to an accuracy of 89.2\%. Notably, PreCrop takes longer than UPaI to train a ResNet-20 model at 807 seconds to a lower accuracy of 88.73\%. Compared to a smaller model, ResNet-8, \Reconv reaches peak accuracy 1.09$\times$ faster. Meanwhile, ResNet-8 outperforms PreCrop in both accuracy and training time.

\textbf{Observation 2}: \Reconv achieves a higher accuracy faster than other SPaI/UPaI methods and smaller models. In fact, \Reconv trains a pruned VGG-16 to within 0.1\% of UPaI 3.37$\times$ faster.

\subsection{Comparison against other SPaI methods}
\label{exp:systems}
In this experiment, \Reconvv, three alternative SPaI methods, and random pruning are evaluated based on accuracy change, GPU speedup, and compression to compare pruned model quality across a range of sparsity values. VGG-19 (CIFAR-10) is the baseline dense model. The three alternative SPaI methods are PreCrop~\cite{cai2022structured}, ProsPr~\cite{prospr}, and 3SP~\cite{3SP}. Table~\ref{tab3} presents the results for all methods\footnote{Some results are missing since the source code for structured pruning used by 3SP and ProsPr are not publicly available, and we could not fully replicate their results. Nonetheless, these results are comparable to \Reconvv.}.

At 80\% sparsity, \Reconv creates the fastest and most compressed pruned model. However, ProsPr maintains 0.09\% more accuracy. At 90\% sparsity, PreCrop obtains a higher compression ratio and speedup, with a 0.14\% lower accuracy than \Reconvv. The same trend is observed at 95\% sparsity. However, the accuracy gap between PreCrop and \Reconv is now 0.41\%. 3SP has the lowest accuracy outside of random pruning for all sparsities except for 95\%, where PreCrop is lower. ProsPr achieves the smallest speedup because it mainly prunes the fully connected layers of VGG-19~\cite{prospr}. Pruning these layers does not decrease inference latency as much as pruning the convolutional layers. At higher sparsity levels, PreCrop outperforms all other methods in terms of speedup and compression because it prunes every layer in the DNN~\cite{cai2022structured}. However, this comes at the expense of reduced model accuracy than other methods.

\textbf{Observation 3}: \Reconv effectively balances the quality of pruned models, making it suitable for edge deployments. It maintains high accuracy while also speeding up and compressing the model significantly.

\begin{table}[t]
    \centering
    \caption{Top-1 accuracy change from the dense baseline model, GPU speedup, and compression for SPaI methods using VGG-19 (CIFAR-10).}
    \begin{tabular}{clrrr}
    \hline
         \scriptsize{Sparsity}&  \scriptsize{Method}&  \scriptsize{Acc. $\Delta$}&  \scriptsize{Speedup ($\times$)}& \scriptsize{Comp. ($\times$)}\\ \hline
         80\%
         &  Random&  \texttt{$-$}1.60&  -& -\\ 
         &  3SP~\cite{3SP}&  \texttt{$-$}0.20&  -& -\\
         &  ProsPr~\cite{prospr}&  \texttt{$+$}0.01&  1.10& -\\
         &  PreCrop~\cite{cai2022structured}& \texttt{$-$}0.07 & 1.35 & 4.55\\
         &  \cellcolor{gray!25}\Reconvv&\cellcolor{gray!25}\texttt{$-$}0.08& \cellcolor{gray!25}1.36& \cellcolor{gray!25}4.66\\
         \hline
         90\%
        &  Random&  \texttt{$-$}3.20&  -& -\\ 
        &  3SP~\cite{3SP} &  \texttt{$-$}0.50&  -& -\\
        &  ProsPr~\cite{prospr} &  0.00&  1.26& -\\
        &  PreCrop~\cite{cai2022structured}& \texttt{$-$}0.26 & 1.63 & 8.89\\
        &  \cellcolor{gray!25}\Reconvv &  \cellcolor{gray!25}\texttt{$-$}0.12&  \cellcolor{gray!25}1.43& \cellcolor{gray!25}5.33\\
        \hline
 95\%
& Random& \texttt{$-$}4.60& -&-\\ 
& 3SP~\cite{3SP}& \texttt{$-$}1.10& -&-\\
& ProsPr~\cite{prospr}& \texttt{$-$}0.28& 1.30&-\\
&  PreCrop~\cite{cai2022structured}& \texttt{$-$}1.22 & 1.79 & 17.59\\
& \cellcolor{gray!25}\Reconvv & \cellcolor{gray!25}\texttt{$-$}0.81& \cellcolor{gray!25}1.44&\cellcolor{gray!25}5.46\\
\hline
    \end{tabular}
    \vspace{-8pt}
    \label{tab3}
\end{table}

\subsection{Overheads}
\label{exp:overheads}
In this experiment, \Reconv is assessed as a low-overhead NAS method for producing pruned models. It is contrasted with other methods that discover their compressed models from an expansive search space. Moreover, \Reconv is compared against traditional pruning after training (PaT). Table~\ref{tab4} presents the results for \Reconv against four NAS methods and $l^{1}$ norm pruning after training (PaT). Search time is defined as the time to find the model candidate. The pruning methods achieve this in one shot by pruning the dense model into a smaller pruned model. NAS methods create many hundreds to thousands of candidate compressed models and then evaluate each for optimality. \Reconv is 2.19$\times$ faster than $l^{1}$ norm since \Reconv prunes based on layer metrics whereas $l^{1}$ norm prunes based on the $l^{1}$ norm of each convolutional filter. In addition, \Reconv produces a pruned model which trains to a higher accuracy than $l^{1}$ norm PaT. Compared to the NAS methods, \Reconv is 1,000$\times$ to 15,000$\times$ faster than NAS at discovering a pruned model and uses up to 38$\times$ less system memory. In addition, the model accuracy is higher than all other NAS methods except for ZenNAS~\cite{zennas}\footnote{ZenNAS has a post-search training regime 10$\times$ longer than typical, which results in a much higher final accuracy than all other methods.}.

\textbf{Observation 4}: \Reconv serves as a search method to identify new pruned models suitable for edge devices, which traditionally relied on their larger cloud counterparts. \Reconv operates significantly faster and yields models of similar accuracy. Its reduced memory requirements further enable its execution on resource-constrained edge devices.

\begin{table}[t]
    \centering
    \caption{Search time and memory usage for various search methods to create a compressed model trained on CIFAR-10. Neural architecture search (NAS) methods target 0.5M parameters. Pruning after training (PaT) and pruning at initialization (PaI) methods prune VGG-16 to a sparsity of 95\% ($p=0.95$, $\sim$0.5M parameters). The top-1 accuracy change is based on a reference VGG-16 (CIFAR-10).}
    \begin{tabular}{lcrrr}
    \hline
         \scriptsize{Method}&  \scriptsize{Type}&  \scriptsize{Search Time (s)}&  \scriptsize{Memory Usage (GB)} & \scriptsize{Acc. $\Delta$}\\ \hline
        ZenNAS~\cite{zennas} & NAS & 19,944 & 4.8 & \texttt{$+$}2.70 \\
        DARTSv2~\cite{darts} & NAS & 1,957 & 8.61 & \texttt{$-$}1.30 \\ 
        NASNet~\cite{nasnet} & NAS & 1,468 & 11.73 & \texttt{$-$}1.32 \\ 
        NASWOT~\cite{naswot} & NAS & 306 & - & \texttt{$-$}0.88\\ 
        $l^{1}$ norm~\cite{classical_struc} & PaT & 3 & 5.12 & \texttt{$-$}2.39 \\ 
        \cellcolor{gray!25}\Reconv & \cellcolor{gray!25}PaI & \cellcolor{gray!25}1.37 & \cellcolor{gray!25}0.31 & \cellcolor{gray!25}\texttt{$-$}0.81 \\ 
\hline
    \end{tabular}
    \vspace{-8pt}
    \label{tab4}
\end{table}

\section{Related Work}
\label{sec:relatedwork}
\textbf{Other Compression Methods} such as quantization reduces the bit precision of DNN parameters to shrink the model and increase inference speed~\cite{quant}. However, it often leads to accuracy loss and might require specialized hardware for lower-precision inference. Knowledge distillation trains a smaller student model using training knowledge from a larger teacher model~\cite{kd}. Although the student can often match the accuracy of the teacher model and take up less space, it is not easily adaptable to different model architectures. Consequently, it does not scale for the heterogeneous edge settings.

\textbf{Pruning Systems} are employed to compress DNN models, creating a range of compact models suited for edge devices. These pruned models optimize resource constraints while maintaining performance on edge deployments. Existing work focuses on pruning after model training~\cite{easiedge, ECCLES2023}, or target model architectures for hardware accelerators\cite{sparselib2}.

\textbf{Neural Architecture Search} (NAS) automates the process of finding optimal DNN model architectures. It is conventionally used to discover larger models that train to higher accuracies~\cite{nasnet}. However, NAS has also been employed to discover smaller models optimized for edge devices~\cite{ofa}. While NAS is effective at discovering high-quality models, repeating the NAS pipeline for a new dataset can be both time-consuming and resource-intensive~\cite{ofa}.

\textbf{Model Reparameterization} aims to optimize model structures, enhancing hardware utilization and thereby improving inference efficiency. For example, RepVGG~\cite{repvgg} reparameterizes ResNet architectures into VGG-style models. However, these methods only work on specific pairs of model architectures and require hardware accelerators, such as GPUs, to maximize utilization.

\section{Conclusions}
\label{sec:conlusion}
Pruning at initialization (PaI) allows for compressed models to be discovered rapidly. However, existing structured PaI methods sacrifice model accuracy to achieve the necessary performance increase required for resource-constrained edge computing. \Reconv addresses this concern by closing the performance gap between unstructured and structured PaI while maintaining the accuracy of unstructured PaI through selective structured pruning of non-sensitive model layers. \Reconv has been shown to work across a range of model architectures and datasets and serves as a foundation for future structured PaI methods.

\section*{Acknowledgments}
This research is funded by Rakuten Mobile, Inc., Japan.

\balance
\bibliographystyle{IEEEtran}  
\bibliography{references}

\end{document}